\documentclass[10pt,twocolumn,letterpaper]{article}

\usepackage{cvpr}              

\usepackage{graphicx}
\usepackage{amsmath}
\usepackage{amssymb}
\usepackage{paralist}
\usepackage{booktabs}       
\usepackage{xcolor}         
\usepackage{multirow}
\usepackage[normalem]{ulem}
\usepackage{float}
\usepackage{makecell}
\usepackage{algorithm}
\usepackage{algpseudocode}
\usepackage{pifont}
\usepackage{bbding}

\usepackage[pagebackref,breaklinks,colorlinks]{hyperref}

\usepackage[capitalize]{cleveref}
\crefname{section}{Sec.}{Secs.}
\Crefname{section}{Section}{Sections}
\Crefname{table}{Table}{Tables}
\crefname{table}{Tab.}{Tabs.}

\begin{document}

\title{Vivid-VR: Distilling Concepts from Text-to-Video Diffusion Transformer for Photorealistic Video Restoration}

\author{
Haoran Bai ~~ Xiaoxu Chen ~~ Canqian Yang ~~ Zongyao He ~~ Sibin Deng ~~ Ying Chen\footnotemark[1]\\
Alibaba Group - Taobao \& Tmall Group\\
\url{https://github.com/csbhr/Vivid-VR}
}

\twocolumn[{
\maketitle
\vspace{-5mm}
\begin{figure}[H]
\hsize=\textwidth
\centering
\includegraphics[width=2.08\linewidth]{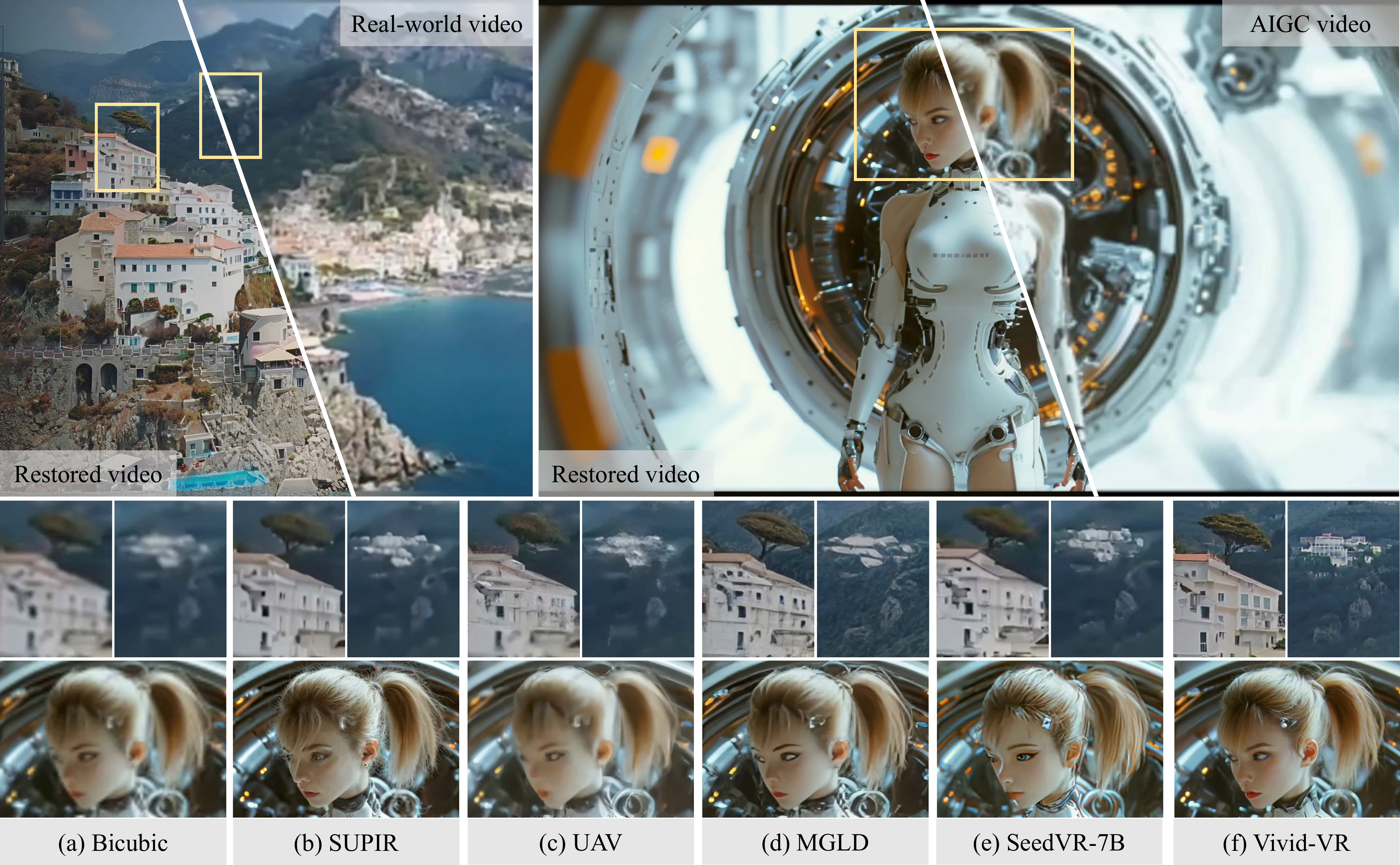}
\vspace{2mm}
\caption{%
Video restoration results on both real-world and AIGC videos. 
To mitigate drift during fine-tuning of the controllable generation pipeline, we propose a concept distillation strategy that preserves both texture realism and temporal coherence in generated videos.
Leveraging this strategy, Vivid-VR achieves impressive texture realism and visual vividness.
(\textbf{Zoom-in for best view})
}
\label{fig:teaser}
\vspace{3mm}
\end{figure}
}]

\renewcommand{\thefootnote}{\fnsymbol{footnote}}
\footnotetext[1]{Corresponding author.}


\begin{abstract}
\vspace{-3mm}

We present Vivid-VR, a DiT-based generative video restoration method built upon an advanced T2V foundation model, where ControlNet is leveraged to control the generation process, ensuring content consistency.
However, conventional fine-tuning of such controllable pipelines frequently suffers from distribution drift due to limitations in imperfect multimodal alignment, resulting in compromised texture realism and temporal coherence. 
To tackle this challenge, we propose a concept distillation training strategy that utilizes the pretrained T2V model to synthesize training samples with embedded textual concepts, thereby distilling its conceptual understanding to preserve texture and temporal quality.
To enhance generation controllability, we redesign the control architecture with two key components: 1) a control feature projector that filters degradation artifacts from input video latents to minimize their propagation through the generation pipeline, and 2) a new ControlNet connector employing a dual-branch design. This connector synergistically combines MLP-based feature mapping with cross-attention mechanism for dynamic control feature retrieval, enabling both content preservation and adaptive control signal modulation.
Extensive experiments show that Vivid-VR performs favorably against existing approaches on both synthetic and real-world benchmarks, as well as AIGC videos, achieving impressive texture realism, visual vividness, and temporal consistency.

\end{abstract}


\vspace{-4mm}
\section{Introduction}
\vspace{-1mm}
\label{sec:intro}

\begin{figure*}[!t]
  \centering
   \includegraphics[width=1.0\linewidth]{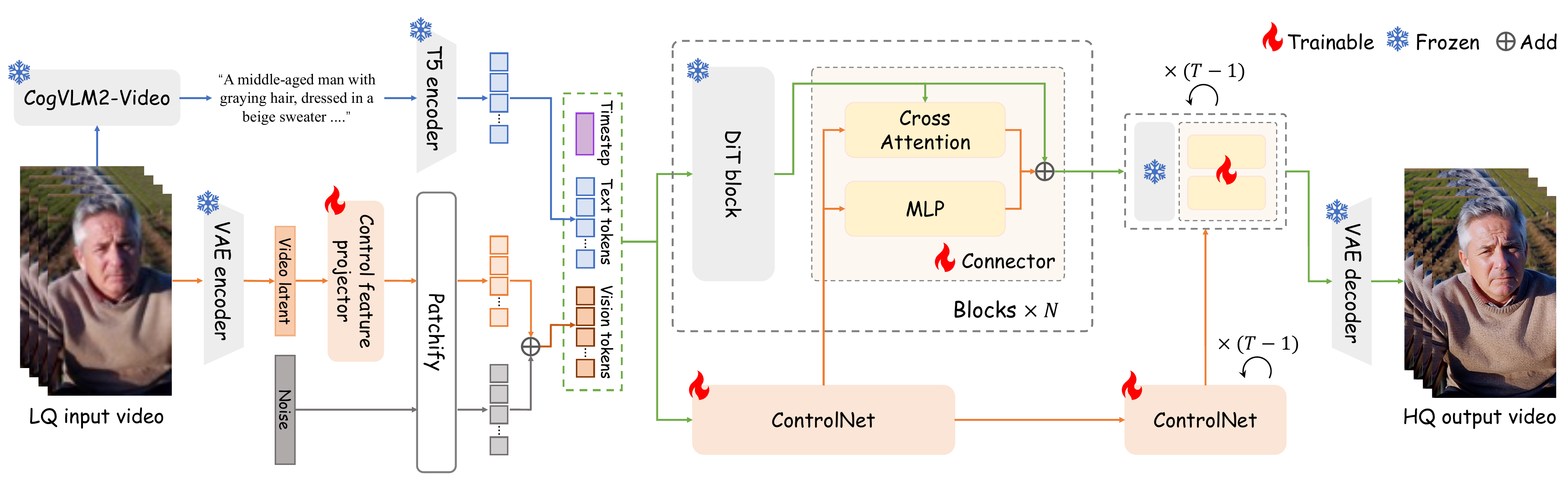}
   \caption{%
    An overview of the proposed method.
    Vivid-VR first processes the LQ input video with CogVLM2-Video to generate a text description, which is encoded into text tokens via T5 encoder.
    Simultaneously, the 3D VAE encoder converts the input video into latent, where our control feature projector removes degradation artifacts.
    The video latent is then patchified, noised, and combined with text tokens and timestep embeddings as input to DiT and ControlNet.
    For enhanced controllability, we introduce a dual-branch connector: an MLP for feature mapping and a cross-attention branch for dynamic control feature retrieval, enabling adaptive input alignment.
    After $T$ denoising steps, the 3D VAE decoder reconstructs the HQ output.
    Only the control feature projector, ControlNet, and connectors are trained via the proposed concept distillation strategy, and other parameters remain frozen.
   }
   \label{fig:framework}
\end{figure*}

Video restoration aims to recover lost textures, fine details, and structural information from low-quality (LQ) input videos to produce high-quality (HQ) ones.
Traditional reconstruction-based methods typically employ CNNs~\cite{wang2019edvr,pan2021deep,chan2021basicvsr,chan2022basicvsr++,chan2022investigating} and Transformers~\cite{liang2024vrt,liang2022recurrent} to extract visual cues for quality enhancement.
However, these approaches face inherent limitations due to insufficient prior knowledge and the ill-posed nature of the inverse problem, reconstructing high-quality textures directly from severely degraded inputs remains extremely challenging.
While GAN-based methods~\cite{wang2018esrgan,wang2021real} can generate some textures to a certain extent, their generative capacity remains limited.

Recent years have witnessed significant advancements in diffusion-based generative models~\cite{rombach2022high,podell2023sdxl,blattmann2023stable}, which can now synthesize photorealistic content. This progress has established generative video restoration as a promising new paradigm.
While initial explorations using text-to-image (T2I) diffusion models have shown impressive results in image restoration tasks~\cite{wang2024exploiting,yu2024scaling,chen2025faithdiff}, their direct application to video sequences suffers from significant temporal inconsistencies due to inadequate motion modeling.
Early attempts to address this limitation typically incorporate temporal enhancement mechanisms, including adding trainable temporal layers to diffusion denoisers and VAE decoders~\cite{zhou2024upscale}, or employing optical flow-based motion compensation~\cite{yang2024motion}.
However, these post-modifications during model fine-tuning are insufficient for achieving robust temporal coherence.
The advent of Diffusion Transformers (DiT)~\cite{peebles2023scalable} has enabled a significant leap forward, with text-to-video (T2V) models~\cite{yang2024cogvideox} now capable of generating both high-quality and temporally stable video content.
This has spurred the development of T2V-based restoration approaches. For instance, SeedVR~\cite{wang2025seedvr} integrates the shift-window attention mechanism with DiT for computational efficiency, and STAR~\cite{xie2025star} proposes a dynamic frequency loss for enhanced fidelity, both achieving decent results.

Despite their advancements, current restoration methods still underperform native T2V models in both texture realism and temporal coherence. This performance gap stems primarily from distribution drift induced by imperfect multimodal alignment during the fine-tuning process.
This issue is not prominent in the T2V model pretraining phase because of the large, diverse training dataset. But the challenge becomes significantly amplified when fine-tuning these models for video restoration, manifesting as unrealistic textures and compromised temporal consistency.

To overcome this challenge, we propose a concept distillation training strategy that leverages synthetic data generated by a pre-trained T2V model.
The proposed approach begins with a source video and its corresponding text description obtained through a vision-language model (VLM).
We first corrupt the source video with noise, then employ the pre-trained T2V model to perform denoising while incorporating the text description.
This process yields a video that encapsulates the T2V model's semantic understanding of the textual concepts, ensuring inherent modal alignment between the generated video and text description in the T2V model's latent space.
By blending these synthesized data with real training samples during fine-tuning, our method successfully transfers the T2V model's conceptual knowledge to the video restoration model, thereby mitigating the distribution drift problem while preserving both texture realism and temporal coherence.

Furthermore, we use ControlNet~\cite{zhang2023adding} for generation control and introduce two key innovations.
First, we develop a control feature projector, which effectively filters degradation artifacts to minimize their propagation through the generation pipeline.
While FaithDiff~\cite{chen2025faithdiff} achieves similar functionality by jointly fine-tuning the VAE encoder which is expensive to train, our solution implements this feature projector as a lightweight CNN-based extension to the VAE encoder.
Second, we redesign the ControlNet connector with a dual-branch architecture.
Different from existing connectors~\cite{yu2024scaling} which fail to properly consider DiT features during fusion, we combine an MLP branch with a cross-attention mechanism, enabling dynamic feature retrieval that preserves the generation quality and realism of native T2V models.
Benefiting from these improvements, the proposed method, named Vivid-VR, achieves impressive texture realism and visual vividness (see Figure~\ref{fig:teaser}).

In summary, our main contributions are as follows:
\begin{compactitem}
	\item We propose a novel concept distillation training strategy that leverages a pre-trained T2V model to synthesize aligned text-video pairs, effectively mitigating distribution drift during fine-tuning and preserving texture and temporal quality.
	\item We improve the ControlNet architecture by introducing a lightweight control feature projector and a dual-branch connector, enabling degradation artifact removal and dynamic control feature retrieval.
	\item The proposed Vivid-VR performs favorably against existing methods on both synthetic and real-world benchmarks, as well as AIGC videos.
\end{compactitem}

\section{Related Work}
\vspace{-1mm}

\noindent\textbf{Reconstruction-based Video Restoration.}
Early approaches focused on the architecture design and loss functions for direct HQ reconstruction from degraded inputs.
CNN-based methods employed various strategies for temporal information integration, including optical flow~\cite{caballero2017real,pan2020cascaded}, deformable convolutions~\cite{wang2019edvr}, bidirectional feature propagation~\cite{chan2021basicvsr}, and optical flow-guided deformable alignment~\cite{chan2022basicvsr++}.
Transformer-based methods~\cite{liang2024vrt,liang2022recurrent} improved performance through attention mechanisms for long-term spatio-temporal modeling.
Meanwhile, some studies~\cite{wang2021real,zhang2021designing} have introduced more complex degradation simulations to improve real-world generalization.
To produce richer textural details, GAN-based frameworks~\cite{wang2018esrgan,wang2021real} are consequently adopted that incorporate adversarial training.
Despite these advances, methods relying solely on input-derived cues without strong priors still produce overly smoothed results when handling severely degraded content.

\noindent\textbf{Diffusion-based Video Restoration.}
In recent years,  diffusion-based generative models~\cite{rombach2022high,podell2023sdxl,blattmann2023stable,yang2024cogvideox} have made significant progress, which introduce a new paradigm for restoration tasks.
Initial explorations focused on image restoration~\cite{wang2024exploiting,yu2024scaling,chen2025faithdiff}.
Based on Stable Diffusion~\cite{rombach2022high}, StableSR~\cite{wang2024exploiting} developed a trainable time-aware encoder to embed input image information into the generation process, transforming it into an image restoration model.
SUPIR~\cite{yu2024scaling} achieved remarkable results through model scaling for image restoration. 
However, these approaches fundamentally lack temporal modeling capabilities, resulting in severe frame inconsistencies when directly applied to video sequences.
Early solutions~\cite{zhou2024upscale,yang2024motion} attempted to mitigate this through temporal enhancement techniques, such as incorporating trainable temporal layers~\cite{zhou2024upscale}, or implementing optical flow-based motion compensation~\cite{yang2024motion}, yet these post-hoc adjustments proved inadequate for ensuring robust temporal coherence.
Diffusion Transformers (DiT)~\cite{peebles2023scalable} enabled high-quality T2V generation~\cite{yang2024cogvideox} with superior temporal stability, inspiring video restoration methods.
SeedVR~\cite{wang2025seedvr} combines the shift-window attention mechanism with DiT to improve computational efficiency.
STAR~\cite{xie2025star} designs a dynamic frequency loss function to improve fidelity.
Concurrently, efforts to improve inference efficiency have led to the design of one-step diffusion models~\cite{wang2025seedvr2,chen2025dove}.
Nevertheless, existing methods still exhibit noticeable gaps in texture realism and temporal consistency compared to native T2V models, due to distribution drift from imperfect multimodal alignment of training data.
To bridge this gap, our work introduces a concept distillation training strategy that effectively preserves the texture and temproal quality of the base T2V model.

\section{Method}
\label{sec:method}

The proposed Vivid-VR leverages the advanced T2V model (i.e., CogVideoX1.5-5B~\cite{yang2024cogvideox}) as its foundation, incorporating the ControlNet~\cite{zhang2023adding} to condition the generation process on input videos.
Figure~\ref{fig:framework} shows an overview of the proposed method.
In this section, we first present the model architecture of the proposed method, and then explain the proposed concept distillation training strategy.

\begin{figure}[!t]
  \centering
   \includegraphics[width=0.99\linewidth]{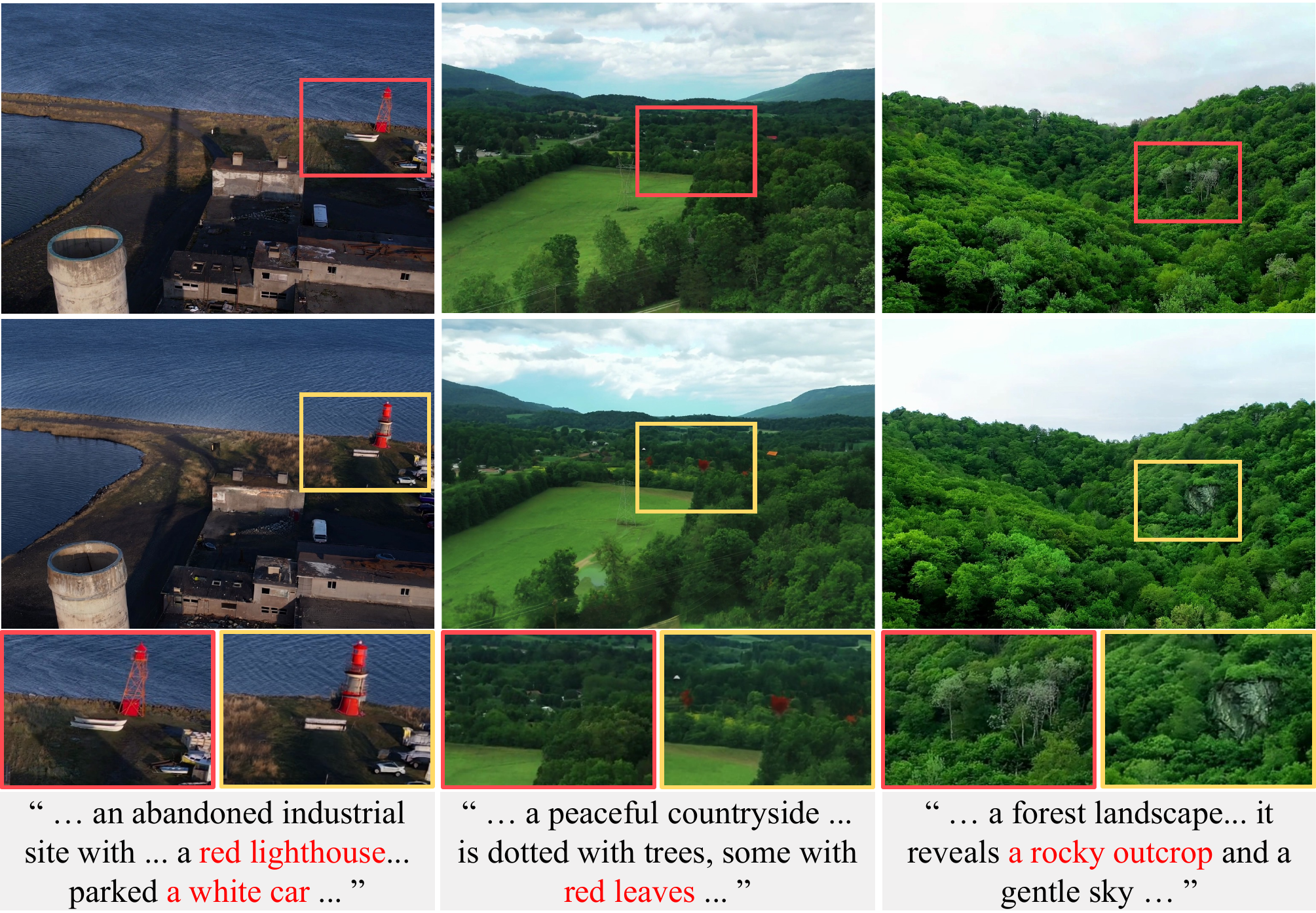}
   \vspace{1mm}
   \caption{%
   	Example videos generated by the proposed concept distillation training strategy.
   The top row presents source videos, and the second row shows corresponding generated videos after embedding textual concepts via the T2V model.
   Due to VLM captioner limitations, the source videos exhibit imperfect alignment with their text descriptions, while the generated videos have better modality alignment. (\textbf{Zoom-in for best view})
   }
   \label{fig:v2v-example}
\end{figure}

\subsection{Model Architecture}
\label{sec:model-architecture}

\noindent\textbf{Text Description Generation.}
Building upon the T2V-based framework, the proposed method requires both low-quality (LQ) input video and corresponding text descriptions.
We employ CogVLM2-Video~\cite{yang2024cogvideox} for text generation to maintain consistency with CogVideoX1.5-5B's training configuration, where this VLM served as the captioner.
Given the input LQ video, CogVLM2-Video produces an aligned text description, subsequently encoded into text tokens through the T5~\cite{raffel2020exploring} text encoder.

\vspace{1mm}
\noindent\textbf{Control Feature Preprocessing.}
In parallel with text tokens generation, we preprocess the LQ input video to generate corresponding visual tokens for DiT and ControlNet.
The preprocessing pipeline begins by encoding LQ video through the VAE encoder, producing the latent representation that contains both content information and degradation artifacts.
Since these degradation artifacts may negatively impact generation quality, we propose a lightweight \textit{Control Feature Projector} to eliminate them.
The proposed projector consists of three cascaded spatiotemporal residual blocks that effectively filter the degraded features, outputting a cleaner latent representation.
The video latent is then patchified and noise injected to form the visual tokens for subsequent processing.

\vspace{1mm}
\noindent\textbf{ControlNet Pipeline.}
Given the text tokens, the visual tokens and the timestep embedding, DiT and ControlNet both perform $T$ denoising steps.
DiT comprises $N$ DiT blocks, while ControlNet contains $N/7$ blocks initialized from DiT's first $N/7$ ones.
During denoising process, ControlNet's visual tokens are integrated into DiT through $N$ proposed \textit{Dual-branch Connectors}.
For the $i^{th}$ connector, the fusion process of the control visual tokens is:
\begin{equation}
	\setlength{\abovedisplayskip}{3pt}  
	\setlength{\belowdisplayskip}{3pt}  
	\hat{f}^{i} = f^{i} + MLP(c^{\lfloor i/7 \rfloor}) + CA(f^{i}, c^{\lfloor i/7 \rfloor}),
	\label{eq:connector}
\end{equation}
where $f^{i}$ denotes the visual tokens from the $i^{th}$ DiT block; $c^{\lfloor i/7 \rfloor}$ represents the corresponding ControlNet block visual tokens aligned with the $i^{th}$ DiT block; $MLP(\cdot)$ and $CA(\cdot)$ are the MLP layer and cross attention module respectively; $\hat{f}^{i}$ is the fused visual tokens.
After $T$ denoising steps, the visual tokens are unpatchified and fed into the VAE decoder to generate the final high-quality (HQ) outputs.

\subsection{Concept Distillation Training Strategy}
\label{sec:concept-distillation}

\noindent\textbf{Training Data Collection.}
Effective training of DiT-based video restoration models demands extensive high-quality text-video pairs, but existing public datasets~\cite{su2017deep,nah2019ntire,xue2019video,stergiou2022adapool} lack in both scale and diversity.
To address this, we collected a large-scale video pool consisting of $3$ million videos with resolutions higher than $1024\times1024$, frame rates higher than $24$, and durations higher than $2$ seconds.
These videos cover a wide range of scenes, including portraits, natural landscapes, plants and animals, urban landscapes, etc.
To ensure video quality, we further screened these videos using the no-reference video quality assessment metrics~\cite{wu2023exploring,zhang2023md} to remove low-quality videos.
For the remaining HQ videos, we generated text descriptions using CogVLM2-Video~\cite{yang2024cogvideox}, maintaining consistency with CogVideoX1.5-5B's configuration.
The final curated multimodal training dataset comprises $500K$ text-video pairs with exceptional quality and variety.

\begin{table*}[!t]
\caption{Quantitative comparisons on benchmarks, including synthetic (SPMCS~\cite{tao2017detail}, UDM10~\cite{yi2019progressive}, YouHQ40~\cite{zhou2024upscale}), real-world (VideoLQ~\cite{chan2022investigating}, UGC50), and AIGC (AIGC50) videos. The best and second performances are marked in {\color[HTML]{F94848} red} and {\color[HTML]{3166FF} blue}, respectively.}
\vspace{1mm}
\renewcommand\arraystretch{1.1}
\footnotesize
\begin{tabular}{c|c|ccccccccc}
\toprule
Datasets    & Metrics    & Real-ESRGAN    & SUPIR    & ~~MGLD~~    & ~~UAV~~    & ~STAR~    & ~DOVE~    & SeedVR-7B    & SeedVR2-7B    & Vivid-VR    \\ \hline
           & PSNR $\uparrow$     & 23.19     & 21.86     & 21.02     & 23.01 & 24.18  & {\color[HTML]{3166FF} 24.80} & 24.08     & {\color[HTML]{F94848} 26.07} & 21.73     \\
           & SSIM $\uparrow$     & 0.690 & 0.609     & 0.595     & 0.606 & 0.720  & {\color[HTML]{3166FF} 0.754} & 0.689     & {\color[HTML]{F94848} 0.777}     & 0.604     \\
           & LPIPS $\downarrow$  & 0.230     & 0.304     & 0.281     & 0.277     & 0.301  & {\color[HTML]{F94848} 0.168} & 0.263 & {\color[HTML]{3166FF} 0.191} & 0.278     \\ \cline{2-11} 
           & NIQE $\downarrow$   & 5.393     & {\color[HTML]{3166FF} 3.494} & 3.790     & 3.503     & 7.058  & 4.031 & 4.514     & 4.969     & {\color[HTML]{F94848} 3.457} \\
           & MUSIQ $\uparrow$    & 51.39     & 65.23     & 58.02     & {\color[HTML]{3166FF} 66.11} & 30.62  & 63.29 & 56.99     & 53.23     & {\color[HTML]{F94848} 70.03} \\
           & CLIP-IQA $\uparrow$ & 0.306     & {\color[HTML]{3166FF} 0.469} & 0.357     & 0.427     & 0.254  & 0.410 & 0.347     & 0.325     & {\color[HTML]{F94848} 0.483} \\
           & DOVER $\uparrow$    & 8.235     & {\color[HTML]{3166FF} 10.07} & 7.981     & 8.987     & 4.266  & 9.898 & 9.779     & 8.625     & {\color[HTML]{F94848} 11.35} \\
\multirow{-9}{*}{SPMCS}       & MD-VQA $\uparrow$   & 79.16     & 82.88 & 78.92     & 81.90     & 74.87  & {\color[HTML]{3166FF} 83.07} & 79.56     & 78.78     & {\color[HTML]{F94848} 86.55} \\ \hline
           & PSNR $\uparrow$     & 27.57     & 27.02     & 28.97 & 28.20 & 27.29  & {\color[HTML]{F94848} 30.53} & 27.80     & {\color[HTML]{3166FF} 29.04}     & 24.54     \\
           & SSIM $\uparrow$     & 0.857 & 0.816     & 0.873 & 0.826     & 0.855  & {\color[HTML]{F94848} 0.894} & 0.848     & {\color[HTML]{3166FF} 0.884}     & 0.761     \\
           & LPIPS $\downarrow$  & 0.187     & 0.208     & 0.158 & 0.196     & 0.167  & {\color[HTML]{F94848} 0.101} & 0.148     & {\color[HTML]{3166FF} 0.117} & 0.243     \\ \cline{2-11} 
           & NIQE $\downarrow$   & 5.835     & {\color[HTML]{3166FF} 4.438} & 4.827     & 5.109     & 6.072  & 5.055 & 5.345     & 5.641     & {\color[HTML]{F94848} 4.046} \\
           & MUSIQ $\uparrow$    & 52.32     & {\color[HTML]{3166FF} 60.84} & 55.82     & 56.19     & 45.38  & 55.17 & 50.29     & 48.91     & {\color[HTML]{F94848} 64.71} \\
           & CLIP-IQA $\uparrow$ & 0.330     & {\color[HTML]{3166FF} 0.418} & 0.339     & 0.333     & 0.289  & 0.340 & 0.273     & 0.272     & {\color[HTML]{F94848} 0.426} \\
           & DOVER $\uparrow$    & 9.402     & {\color[HTML]{3166FF} 10.49} & 9.319     & 9.774     & 9.454  & 10.41 & 9.349     & 8.752     & {\color[HTML]{F94848} 11.97} \\
\multirow{-9}{*}{UDM10}       & MD-VQA $\uparrow$   & 83.51     & {\color[HTML]{3166FF} 85.21} & 83.89     & 83.14     & 82.10  & 83.99 & 80.15     & 79.88     & {\color[HTML]{F94848} 90.05} \\ \hline
           & PSNR $\uparrow$     & 23.02     & 21.57     & 23.24 & 22.31     & 22.92  & {\color[HTML]{F94848} 24.10} & 22.46     & {\color[HTML]{3166FF} 24.00} & 21.31     \\
           & SSIM $\uparrow$     & 0.655 & 0.585     & 0.639     & 0.592     & 0.657  & {\color[HTML]{3166FF} 0.688} & 0.621    & {\color[HTML]{F94848} 0.693} & 0.579     \\
           & LPIPS $\downarrow$  & 0.341     & 0.347     & 0.350     & 0.340     & 0.433  & 0.283 & {\color[HTML]{3166FF} 0.240} & {\color[HTML]{F94848} 0.185} & 0.357     \\ \cline{2-11} 
           & NIQE $\downarrow$   & 4.316     & {\color[HTML]{3166FF} 3.299} & 4.038     & {\color[HTML]{F94848} 3.127} & 6.744  & 4.456 & 4.243     & 4.576     & 3.410     \\
           & MUSIQ $\uparrow$    & 60.03     & {\color[HTML]{3166FF} 68.46}     & 59.40     & 65.97     & 36.36  & 60.65 & 61.91     & 59.34     & {\color[HTML]{F94848} 70.55} \\
           & CLIP-IQA $\uparrow$ & 0.389     & {\color[HTML]{F94848} 0.485} & 0.362     & 0.427     & 0.279  & 0.356 & 0.360     & 0.336     & {\color[HTML]{3166FF} 0.447}     \\
           & DOVER $\uparrow$    & 12.60     & 12.93     & 11.01     & 12.36     & 7.868  & 12.52 & {\color[HTML]{3166FF} 14.00} & 12.80     & {\color[HTML]{F94848} 14.61} \\
\multirow{-9}{*}{YouHQ40}     & MD-VQA $\uparrow$   & 88.85     & {\color[HTML]{3166FF} 89.44} & 86.24     & 87.35     & 76.89  & 86.51 & 87.51     & 85.82     & {\color[HTML]{F94848} 92.92} \\ \hline
           & NIQE $\downarrow$   & 5.014     & 4.628     & {\color[HTML]{3166FF} 4.565} & 4.591     & 5.789  & 5.049 & 4.994     & 5.674     & {\color[HTML]{F94848} 4.371} \\
           & MUSIQ $\uparrow$    & 55.29     & 54.45     & {\color[HTML]{3166FF} 57.70} & 55.82     & 50.52  & 55.11 & 46.49     & 43.41     & {\color[HTML]{F94848} 62.47} \\
           & CLIP-IQA $\uparrow$ & 0.287     & {\color[HTML]{3166FF} 0.299} & 0.297     & 0.262     & 0.265  & 0.271 & 0.229     & 0.220     & {\color[HTML]{F94848} 0.338} \\
           & DOVER $\uparrow$    & 8.453     & 8.609     & {\color[HTML]{3166FF} 8.830} & 7.777     & 8.758  & 8.780 & 7.240     & 6.331     & {\color[HTML]{F94848} 9.743} \\
\multirow{-5}{*}{VideoLQ}     & MD-VQA $\uparrow$   & {\color[HTML]{3166FF} 80.50} & 77.32     & 80.67     & 78.02     & 78.56  & 79.33 & 74.80     & 73.52     & {\color[HTML]{F94848} 83.14} \\ \hline
           & NIQE $\downarrow$   & 5.866     & 5.396     & {\color[HTML]{3166FF} 4.633} & 5.350     & 5.754  & 5.493 & 5.662     & 6.230     & {\color[HTML]{F94848} 4.361} \\
           & MUSIQ $\uparrow$    & 52.22     & 58.25     & {\color[HTML]{3166FF} 61.42} & 54.71     & 55.01  & 57.82 & 49.76     & 46.12     & {\color[HTML]{F94848} 67.61} \\
           & CLIP-IQA $\uparrow$ & 0.318     & 0.382     & {\color[HTML]{3166FF} 0.396} & 0.353     & 0.353  & 0.353 & 0.305     & 0.276     & {\color[HTML]{F94848} 0.450} \\
           & DOVER $\uparrow$    & 10.25     & {\color[HTML]{3166FF} 12.01} & 11.78     & 10.44     & 10.92  & 11.84 & 10.47     & 8.209     & {\color[HTML]{F94848} 14.46} \\
\multirow{-5}{*}{UGC50}       & MD-VQA $\uparrow$   & 80.85     & 82.27     & {\color[HTML]{3166FF} 84.81} & 81.12     & 81.93  & 82.30 & 78.69     & 75.49     & {\color[HTML]{F94848} 88.89} \\ \hline
           & NIQE $\downarrow$   & 5.680     & 5.206     & {\color[HTML]{3166FF} 4.953} & 5.579     & 5.737     & 5.278  & 5.029 & 5.973     & {\color[HTML]{F94848} 4.184} \\
           & MUSIQ $\uparrow$    & 54.26     & 58.11     & 61.39 & 57.62     & 51.66     & {\color[HTML]{3166FF} 62.07}  & 61.61 & 49.35     & {\color[HTML]{F94848} 67.18} \\
           & CLIP-IQA $\uparrow$ & 0.349     & 0.380     & {\color[HTML]{3166FF} 0.391} & 0.376     & 0.309     & 0.379  & 0.378 & 0.290     & {\color[HTML]{F94848} 0.445} \\
           & DOVER $\uparrow$    & 12.36     & 13.33 & 12.70     & 12.28     & 12.10     & {\color[HTML]{3166FF} 14.49}  & 14.46 & 11.34     & {\color[HTML]{F94848} 14.51} \\
\multirow{-5}{*}{AIGC50}      & MD-VQA $\uparrow$   & 84.56     & 84.80     & 85.45 & 83.06     & {\color[HTML]{3166FF} 86.97}  & 85.54 & 81.47     & 80.37     & {\color[HTML]{F94848} 89.69} \\ 
\bottomrule
\end{tabular}
\label{tab:quantitative-results}
\end{table*}

\vspace{1mm}
\noindent\textbf{Concept Distillation.}
Due to limitations of the VLM captioner, the constructed text-video data pairs are not perfectly aligned (see Figure~\ref{fig:v2v-example} top row).
This may lead to distribution drift during fine-tuning, degrading output video quality.
While developing a more accurate VLM captioner could enhance text-video alignment, it has two drawbacks: (1) it is costly, and (2) discrepancies may persist in the T2V model's latent space, potentially still leading to distribution drift.
Instead, we address this issue by distilling the text-concept understanding capabilities of the T2V base model into the video restoration model.
To this end, we employ the T2V model itself (i.e., CogVideoX1.5-5B) to perform text-guided video-to-video translation, generating training data for distillation.
Specifically, given a text-video pair, we perturb the source video by adding noise with a standard deviation corresponding to $T/2$ time steps.
We then apply CogVideoX1.5-5B to denoise the video over $T/2$ steps, conditioned on the text description, yielding a synthesized video with inherent alignment to text concepts in this T2V model's latent space.
As illustrated in Figure~\ref{fig:v2v-example} (second row), the generated video largely retains the source content, but some concepts have been modified to better align with those in the text description.
We randomly extract text-video pairs from the collected training dataset and employ the aforementioned process to generate $100K$ sample pairs.
These generated samples are then combined with the original training dataset to facilitate fine-tuning of the control module in our DiT-based video restoration model.

\vspace{1mm}
\noindent\textbf{Model Training.}
Following the settings of CogVideoX1.5-5B~\cite{yang2024cogvideox}, we employ $v$-prediction for training, and the loss function is:
\begin{equation}
	\setlength{\abovedisplayskip}{3pt}  
	\setlength{\belowdisplayskip}{3pt}  
	\mathcal{L}=\mathbb{E}_{x_0,t,\epsilon} \left [ \left \| v-v_\theta(x_t, x^{lq}, x^{text}, t ) \right \|^2  \right ] ,
	\label{eq:loss}
\end{equation}
where $x_0$ is the HQ video sampled from training dataset;
$x^{text}$ and $x^{lq}$ are the corresponding text description and the synthesized LQ video using the degradation model from~\cite{wang2021real};
$t$ and $\epsilon$ are the time step and noise;
$x_t=\sqrt{\bar{\alpha_t}}x_0 + \sqrt{1-\bar{\alpha_t}}\epsilon$ is the noised latent of $x_0$, and $\bar{\alpha_t}$ is the cumulative multiplication of the variance corresponding to time step $t$;
$v_\theta$ denotes the networks of DiT and ControlNet (including the control feature projector and the connectors);
$v$ is the optimization target, which is defined as $v=\sqrt{\bar{\alpha_t}}\epsilon - \sqrt{1-\bar{\alpha_t}}x_0$.
During training, only the control feature projector, ControlNet, and connectors are trained, and other parameters remain frozen.


\begin{figure*}[!t]
  \centering
   \includegraphics[width=1.0\linewidth]{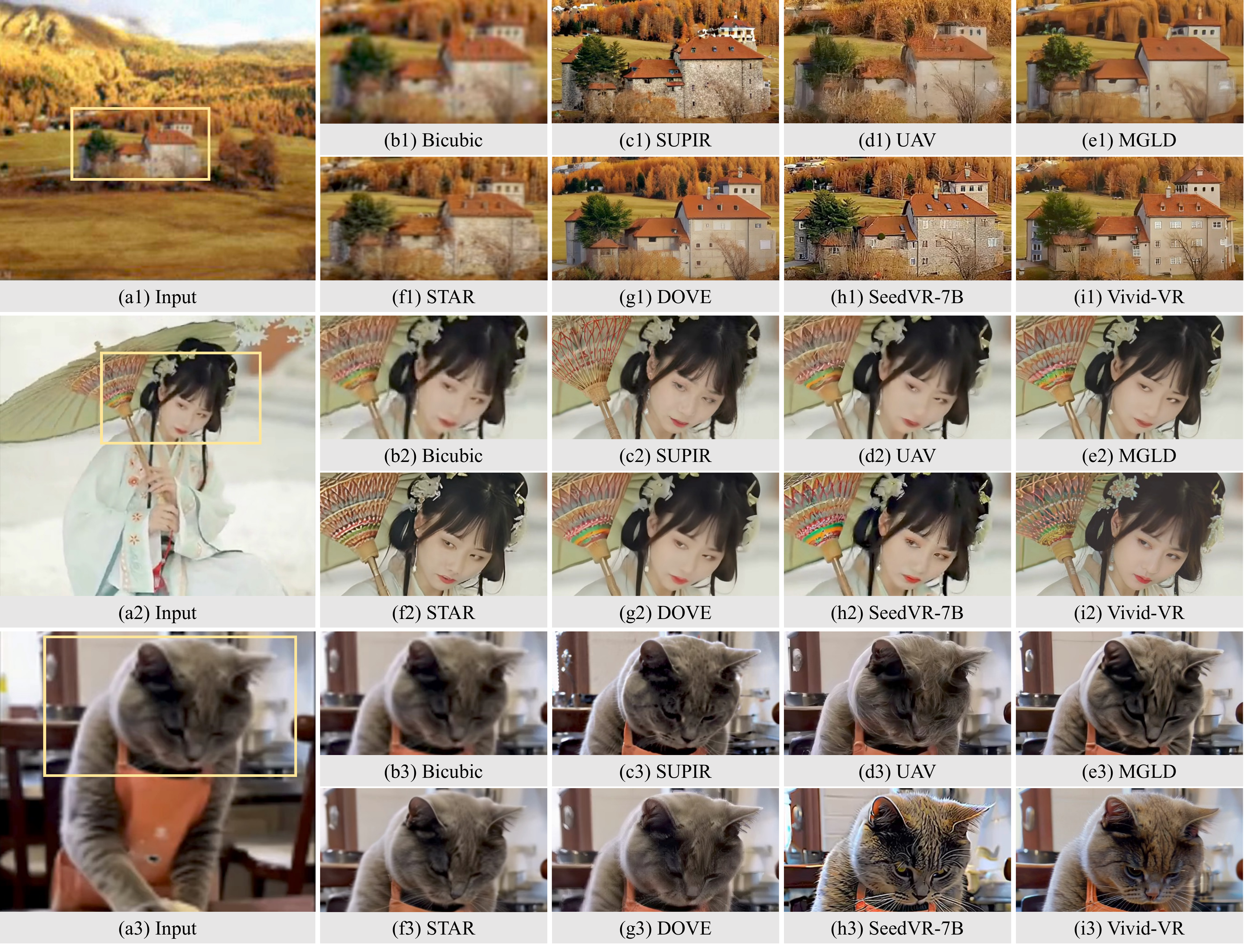}
   \caption{%
    Qualitative comparison results on synthetic (first row), real-world (second row), and AIGC (third row) videos. The proposed Vivid-VR produces the frames with more reasonable structures, as well as more realistic and vivid textures. (\textbf{Zoom-in for best view})
   }
   \label{fig:visual_compare}
\end{figure*}

\vspace{-1mm}
\section{Experimental Results}
\label{sec:experimental-results}
\vspace{-2mm}

In this section, we evaluate the proposed Vivid-VR on both synthetic and real-world benchmark datasets and compare it with state-of-the-art methods.

\subsection{Implementation Details}
\label{sec:implementation-details}

The overall training dataset includes $500K$ real videos and $100K$ generated videos, as well as their corresponding text descriptions.
During training, we resize the short side of these videos to $1024$ pixels and then center-crop them to $1024\times1024$ resolution.
The number of training video frames is randomly selected between $17$ and $37$.
We use the AdamW optimizer~\cite{loshchilov2017decoupled} with a learning rate of $0.0001$, and adopt cosine annealing learning rate scheme~\cite{wang2019edvr}.
We train Vivid-VR on $32$ NVIDIA H20-96G GPUs, with a batch size of $1$ per GPU.
The number of training iterations is $30K$, and the entire training process takes approximately $6K$ GPU hours.
For inference, we set the number of denoising steps to $50$ and used the DPM solver~\cite{lu2025dpm}.
To maintain consistency with training settings, we run inference on videos at $1024\times1024$ resolution.
For higher resolution inputs, we employ aggregation sampling~\cite{wang2024exploiting} with direct block concatenation rather than Gaussian-weighted averaging to prevent overlapping region artifacts.

\begin{figure*}[!t]
	\centering
	\includegraphics[width=1.0\linewidth]{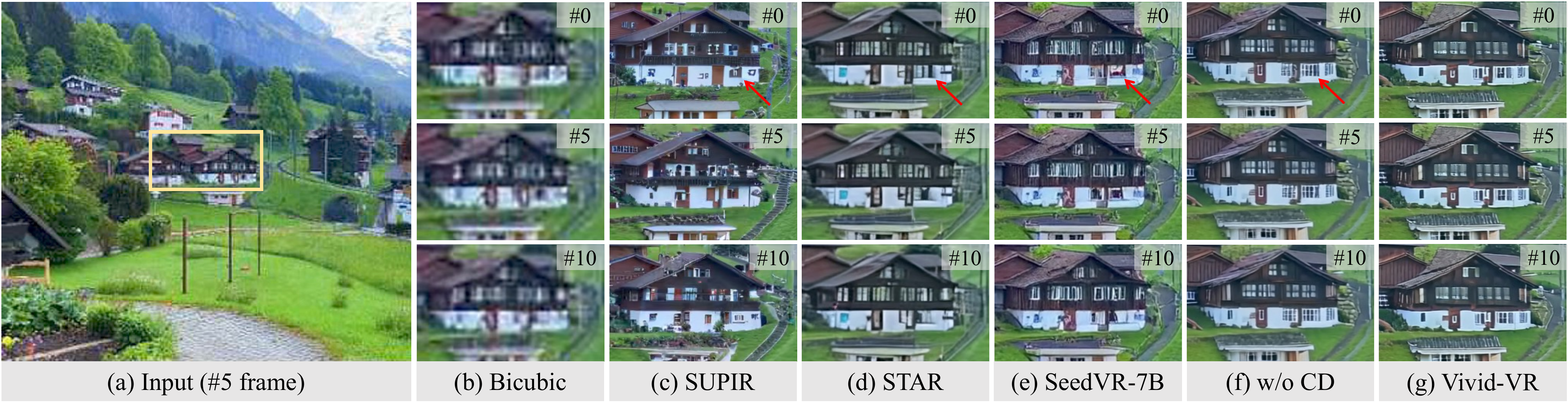}
	
	\vspace{1mm}
	\caption{%
		Visual comparison results on temporal consistency.
		(a) displays the $\#5$ frame of the input video, and (b)-(g) present the outputs at frames $0$, $5$, and $10$, where ``CD" denotes the proposed concept distillation.
		Vivid-VR demonstrates superior temporal coherence, as evident from the consistent structure of windows and doors throughout the sequence.
		(\textbf{Zoom in on the red arrow area in each frame})
	}
	\label{fig:visual_compare_temporal}
\end{figure*}

\begin{table*}[!t]
\caption{Ablation studies of the proposed method on the UGC50 testset, where ``FT" denotes ``fine-tuning", ``CA" denotes ``cross attention branch", ``SK" denotes ``replacing MLP with a skip connection", ``QW" denotes ``Qwen2.5-VL as VLM captioner", and (h) is the setting of the proposed Vivid-VR. The best performances is marked in \textbf{bold} text.}
\renewcommand\arraystretch{1.1}
\footnotesize
\begin{tabular}{c|c|c|c|c|c|c|cccc}
\toprule
\multirow{2}{*}{Methods} & \multicolumn{2}{c|}{{Control Feature Preprocessing}} & \multicolumn{3}{c|}{ControlNet Connectors} & \multirow{2}{*}{Concept Distillation} & \multirow{2}{*}{NIQE$\downarrow$} & \multirow{2}{*}{MUSIQ$\uparrow$} & \multirow{2}{*}{CLIP-IQA$\uparrow$} & \multirow{2}{*}{DOVER$\uparrow$} \\ \cline{2-6}
        &  FT VAE Enc   &  Projector      & ZeroSFT    & MLP    & CA   &      &      &       &          &       \\ \hline
(a)     & \ding{55}  & \ding{55}        & \ding{55}             & \Checkmark       & \Checkmark        & \Checkmark             & 4.622                 & 63.06    & 0.414    & 13.98    \\
(b)     & \Checkmark  & \ding{55}        & \ding{55}             & \Checkmark       & \Checkmark        & \Checkmark             & 4.632                 & 64.31    & 0.408    & 14.40    \\
(c)     & \ding{55}  & \Checkmark        & \ding{55}             & \Checkmark       & \ding{55}        & \Checkmark             & 5.183                 & 59.78    & 0.374    & 13.04    \\
(d)     & \ding{55}  & \Checkmark        & \ding{55}             & \textit{SK}       & \Checkmark        & \Checkmark             &   4.730               &  63.91   &  0.401   &  13.71   \\
(e)     & \ding{55}  & \Checkmark        & \Checkmark             & \ding{55}       & \ding{55}        & \Checkmark             & 4.771                 & 61.21    & 0.389    & 13.77    \\
(f)     & \ding{55}  & \Checkmark        & \ding{55}             & \Checkmark       & \Checkmark        & \ding{55}             & 5.364                 & 57.36    & 0.363    & 12.99    \\
(g)     & \ding{55}  & \Checkmark        & \ding{55}             & \Checkmark       & \Checkmark        & \textit{QW}             &  5.253                &  60.88   & 0.354   &  13.45   \\
(h)     & \ding{55}  & \Checkmark        & \ding{55}             & \Checkmark       & \Checkmark        & \Checkmark             & \textbf{4.361}                 &\textbf{67.61}     & \textbf{0.450}     & \textbf{14.46}    \\
\bottomrule
\end{tabular}
\label{tab:ablation-study}
\end{table*}

\vspace{-1mm}
\subsection{Quantitative Results}
\label{sec:quantitative-results}
\vspace{-2mm}

To evaluate the performance of the proposed algorithm, we compare Vivid-VR against state-of-the-art approaches, including reconstruction-based methods (Real-ESRGAN~\cite{wang2021real}), generative image restoration methods (SUPIR~\cite{yu2024scaling}), and generative video restoration methods (UAV~\cite{zhou2024upscale}, MGLD~\cite{yang2024motion}, STAR~\cite{xie2025star}, DOVE~\cite{chen2025dove}, SeedVR-7B~\cite{wang2025seedvr}, SeedVR2-7B~\cite{wang2025seedvr2}).
The evaluation covers synthetic (SPMCS~\cite{tao2017detail}, UDM10~\cite{yi2019progressive}, YouHQ40~\cite{zhou2024upscale}) and real-world (VideoLQ~\cite{chan2022investigating}) benchmarks.
Furthermore, we construct two testsets, containing real-world UGC videos (UGC50) and AIGC videos (AIGC50).
For real-world and AIGC videos lacking of ground truth, we employed no-reference image (NIQE~\cite{mittal2012making}, MUSIQ~\cite{ke2021musiq}, CLIP-IQA~\cite{wang2023exploring}) and video quality assessments (DOVER~\cite{wu2023exploring}, MD-VQA~\cite{zhang2023md}).
For synthetic benchmarks, we supplemented these no-reference metrics with full-reference evaluations (PSNR, SSIM, and LPIPS~\cite{zhang2018unreasonable}).

Table~\ref{tab:quantitative-results} presents quantitative comparisons on $6$ benchmark testsets.
The proposed Vivid-VR significantly outperforms existing methods in no-reference metrics, achieving the best results in almost all metrics.
At the same time, we also note its advantages in full-reference metrics appear less pronounced.
We argue that this arises primarily from  the inherent limitations of these metrics, which often fail to align with human perceptual preferences.
For example, the LPIPS values of Figure~\ref{fig:visual_compare}(g1) and (i1) are $0.3112$ and $0.4297$, respectively, while Figure~\ref{fig:visual_compare}(i1) is more prefered by human.
This phenomenon becomes particularly evident in generative restoration scenarios where severe input degradation allows multiple plausible HQ outputs, making full-reference metrics inadequate for quality assessment.
This has also been mentioned in~\cite{yu2024scaling} and has been noticed by quality assessment studies~\cite{blau2018perception,jinjin2020pipal,gu2022ntire}.

\subsection{Qualitative Results}
\label{sec:qualitative-results}
\vspace{-1mm}

Figures~\ref{fig:teaser} and~\ref{fig:visual_compare} present visual comparisons with existing methods on synthetic, real-world, and AIGC videos.
The proposed Vivid-VR achieves remarkable texture realism and visual vividness.
Notably, Vivid-VR is able to generate reasonable and clear structures, such as the house shown in Figure~\ref{fig:visual_compare}(i1), while existing methods exhibit structural distortions, artifacts, and loss of fine details (see Figure~\ref{fig:visual_compare}(c1)-(h1)).
Moreover, the proposed Vivid-VR produces more realistic and delicate textures on human portraits and animal fur (see Figure~\ref{fig:visual_compare}(i2) and (i3)), while existing methods frequently yield either overly blurred or oversharpened outputs that are perceptually unrealistic.

In addition, Figure~\ref{fig:visual_compare_temporal} shows visual comparisons on temporal consistency.
As shown in Figure~\ref{fig:visual_compare_temporal}(g), the proposed Vivid-VR demonstrates superior coherence.
For example, the structures of the windows and doors are well-consistent throughout the sequence.
In contrast, SUPIR~\cite{yu2024scaling} shows frame-wise inconsistency as it is an image-based restoration approach (see Figure~\ref{fig:visual_compare_temporal}(c)).
While STAR~\cite{xie2025star} and SeedVR-7B~\cite{wang2025seedvr} leverage T2V frameworks, their fine-tuning-induced distribution drift compromises temporal consistency (see Figure~\ref{fig:visual_compare_temporal}(d)-(e)).
Notably, Vivid-VR exhibits similar degradation when without using the proposed concept distillation strategy (see Figure~\ref{fig:visual_compare_temporal}(f)).

\vspace{-1mm}
\section{Analysis and Discussions}
\label{sec:analysis-discussion}
\vspace{-1mm}

We have shown that the proposed Vivid-VR performs favorably against state-of-the-art methods. 
In this section, we perform further analysis on the key components.

\vspace{-1mm}
\subsection{Effect of the Control Feature Projector}
\label{sec:ablation-control-feature-projector}
\vspace{-1mm}

The T2V base model is trained on HQ videos, making direct use of latent features from LQ videos detrimental to generation quality.
To address this issue, we propose a control feature projector to remove degradation artifacts from the LQ video latents.
To validate its effectiveness, we conducted an ablation study by disabling this module while keeping other settings unchanged. As shown in Table~\ref{tab:ablation-study} ((a) vs (h)), the proposed control feature projector is able to improve video restoration.
SUPIR~\cite{yu2024scaling} attempts to tackle this challenge by independently fine-tuning the VAE encoder.
However, this decoupled optimization creates feature incompatibility with subsequent DiT and ControlNet, leading to suboptimal results (Table~\ref{tab:ablation-study} (b)). 
While this problem can be solved through joint VAE encoder and video restoration optimization~\cite{chen2025faithdiff}, this approach is expensive to train.
In contrast, our proposed lightweight control feature projector can achieve similar effects at lower training cost.

\begin{figure}[!t]
	\centering
	\includegraphics[width=0.99\linewidth]{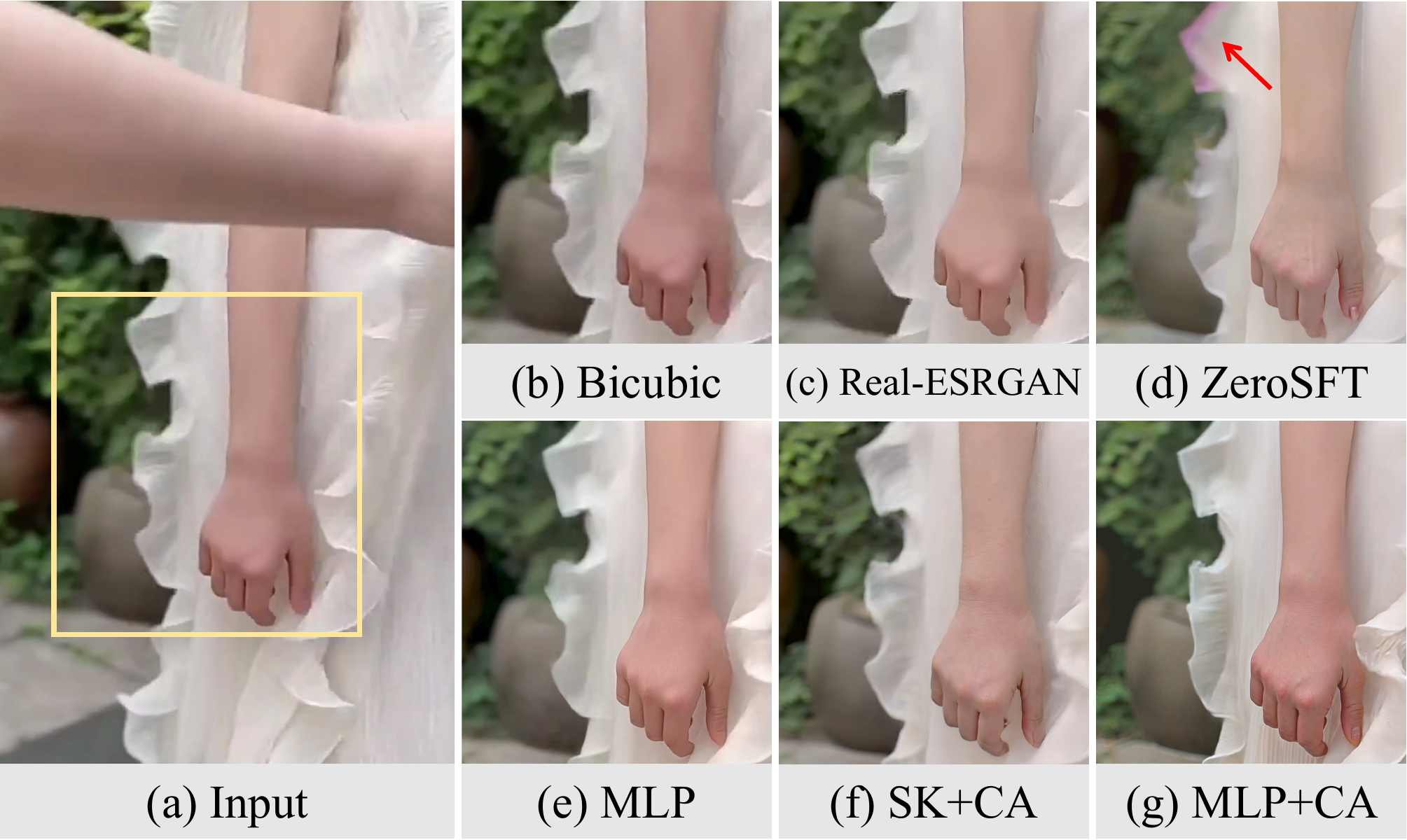}
	\vspace{1mm}
	\caption{%
		Effect of the dual-branch connector, where ``SK" denotes ``replacing MLP with a skip connection", and ``CA" denotes ``cross attention branch". (\textbf{Zoom-in for best view})
	}
	\label{fig:ablation-connectors}
\end{figure}

\begin{figure}[!t]
	\centering
	\includegraphics[width=0.99\linewidth]{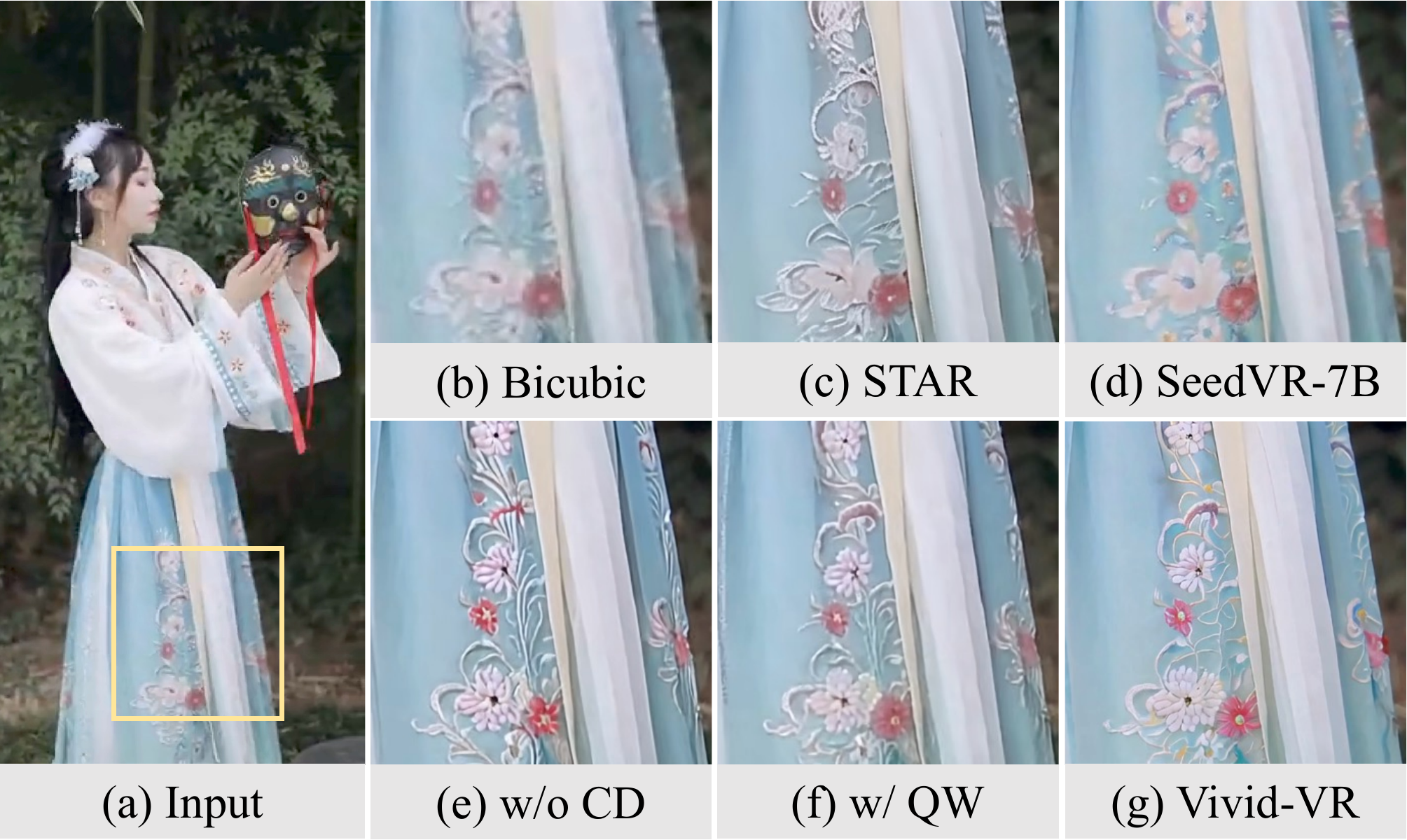}
	\vspace{1mm}
	\caption{%
		Effect of the concept distillation, where ``CD" denotes ``concept distillation strategy", and ``QW" denotes ``Qwen2.5-VL as VLM captioner". (\textbf{Zoom-in for best view})
	}
	\label{fig:ablation-v2v}
\end{figure}

\subsection{Effect of the Dual-Branch Connector}
\label{sec:ablation-connectors}
\vspace{-1mm}

For the ControlNet connector, we propose a dual-branch architecture combining an MLP for feature mapping with a cross attention mechanism for dynamic feature retrieval.
One may wonder whether this design helps video restoration.
To answer this question, we conduct three ablation studies: 1) disabling the cross attention branch; 2) replacing the MLP branch with a skip connection; 3) adopting the ZeroSFT connector~\cite{yu2024scaling}. 
Table~\ref{tab:ablation-study} and Figure~\ref{fig:ablation-connectors} show the comparative results of the ablation experiments.
When the cross attention branch is disabled, the MLP connector does not perform well and produces results lacking in detail (see Table~\ref{tab:ablation-study}(c) and Figure~\ref{fig:ablation-connectors}(e)).
When the MLP branch is simply disabled, the video restoration model fails to converge due to its exclusive selection of DiT-like features from control inputs, resulting in output results that do not match the input content.
To ensure model convergence, we replace the MLP with a skip connection. The results in Table~\ref{tab:ablation-study}(d) and Figure~\ref{fig:ablation-connectors}(f) show that without the MLP feature mapping, the recovered results are not well.
These experiments demonstrate the necessity of our dual-branch design.

In addition, the results in Table~\ref{tab:ablation-study}(e) show that the performance of ZeroSFT connector~\cite{yu2024scaling} is inferior to our proposed dual-branch connector.
Furthermore, the normalization operation in ZeroSFT architecture often causes residual artifacts of adjacent frames to appear in the output frames (see Figure~\ref{fig:ablation-connectors}(d)), while removing the normalization leads to gradient explosion during training.
In contrast, our proposed connector avoids these problems.

\subsection{Effect of the Concept Distillation Strategy}
\label{sec:ablation-concept-distillation}
\vspace{-1mm}

To mitigate distribution drift caused by imperfect multimodal alignment in training data, we introduce a concept distillation strategy, that leverages the T2V base model to generate training data.
To verify its effectiveness, we disable the generated training data and train this baseline method only on the collected videos.
Table~\ref{tab:ablation-study}(f) and Figure~\ref{fig:ablation-v2v}(e) show that the baseline method without the concept distillation strategy fails to achieve high-quality results, showing overly sharp textures due to distribution drift.
For the similar reason, textures generated by STAR and SeedVR are also less realistic.
In addition, the baseline method without the concept distillation also suffers from a decline in temporal coherence (see Figure~\ref{fig:visual_compare_temporal}(f)).
This verifies that our concept distillation can facilitate video restoration in terms of both perceptual quality and temporal consistency.

Furthermore, to verify whether using a more accurate VLM captioner could resolve the distribution drift problem, 
we conduct an ablation study:  using the more advanced Qwen2.5-VL as VLM captioner for training data annotation.
As evidenced by Table~\ref{tab:ablation-study}(g) and Figure~\ref{fig:ablation-v2v}(f), the more accurate VLM captioner (i.e., Qwen2.5-VL) also fails to completely eliminate the modality gap in the T2V model's latent space, demonstrating the persistence of distribution drift even with superior captioning models.

\vspace{-1mm}
\section{Conclusions and Limitations}
\label{sec:conclusions-limitations}
\vspace{-1mm}

We have proposed Vivid-VR, a DiT-based generative video restoration method built upon an advanced T2V foundation model.
To mitigate distribution drift during fine-tuning, we introduced a concept distillation strategy that leverages the pre-trained T2V model to synthesize training samples with embedded textual concepts.
Regarding the model architecture for controllable generation, we proposed two key components: 1) a control feature projector for degradation artifacts removal, and 2) a dual-branch connector combining an MLP and cross-attention mechanism for control feature mapping and dynamic retrieval.
Both quantitative and qualitative experimental results demonstrate the effectiveness of the proposed method.

The proposed method builds upon the CogVideoX1.5-5B T2V model and inherits its inference complexity, which results in lengthy inference times.
Future work will explore ways to enhance the algorithm’s efficiency, such as applying one-step diffusion fine-tuning to achieve comparable video restoration quality in a single forward pass.

\clearpage
{\small
\bibliographystyle{ieee_fullname}
\bibliography{egbib}

\begin{thebibliography}{10}\itemsep=-1pt

\bibitem{blattmann2023stable}
Andreas Blattmann, Tim Dockhorn, Sumith Kulal, Daniel Mendelevitch, Maciej
  Kilian, Dominik Lorenz, Yam Levi, Zion English, Vikram Voleti, Adam Letts,
  et~al.
\newblock Stable video diffusion: Scaling latent video diffusion models to
  large datasets.
\newblock {\em arXiv preprint arXiv:2311.15127}, 2023.

\bibitem{blau2018perception}
Yochai Blau and Tomer Michaeli.
\newblock The perception-distortion tradeoff.
\newblock In {\em CVPR}, pages 6228--6237, 2018.

\bibitem{caballero2017real}
Jose Caballero, Christian Ledig, Andrew Aitken, Alejandro Acosta, Johannes
  Totz, Zehan Wang, and Wenzhe Shi.
\newblock Real-time video super-resolution with spatio-temporal networks and
  motion compensation.
\newblock In {\em CVPR}, pages 4778--4787, 2017.

\bibitem{chan2021basicvsr}
Kelvin~CK Chan, Xintao Wang, Ke Yu, Chao Dong, and Chen~Change Loy.
\newblock Basicvsr: The search for essential components in video
  super-resolution and beyond.
\newblock In {\em CVPR}, pages 4947--4956, 2021.

\bibitem{chan2022basicvsr++}
Kelvin~CK Chan, Shangchen Zhou, Xiangyu Xu, and Chen~Change Loy.
\newblock Basicvsr++: Improving video super-resolution with enhanced
  propagation and alignment.
\newblock In {\em CVPR}, pages 5972--5981, 2022.

\bibitem{chan2022investigating}
Kelvin~CK Chan, Shangchen Zhou, Xiangyu Xu, and Chen~Change Loy.
\newblock Investigating tradeoffs in real-world video super-resolution.
\newblock In {\em CVPR}, pages 5962--5971, 2022.

\bibitem{chen2025faithdiff}
Junyang Chen, Jinshan Pan, and Jiangxin Dong.
\newblock Faithdiff: Unleashing diffusion priors for faithful image
  super-resolution.
\newblock In {\em CVPR}, pages 28188--28197, 2025.

\bibitem{chen2025dove}
Zheng Chen, Zichen Zou, Kewei Zhang, Xiongfei Su, Xin Yuan, Yong Guo, and Yulun
  Zhang.
\newblock Dove: Efficient one-step diffusion model for real-world video
  super-resolution.
\newblock {\em arXiv preprint arXiv:2505.16239}, 2025.

\bibitem{deng2025acquire}
Junyuan Deng, Xinyi Wu, Yongxing Yang, Congchao Zhu, Song Wang, and Zhenyao Wu.
\newblock Acquire and then adapt: Squeezing out text-to-image model for image
  restoration.
\newblock In {\em CVPR}, pages 23195--23206, 2025.

\bibitem{gu2022ntire}
Jinjin Gu, Haoming Cai, Chao Dong, Jimmy~S Ren, Radu Timofte, Yuan Gong,
  Shanshan Lao, Shuwei Shi, Jiahao Wang, Sidi Yang, et~al.
\newblock Ntire 2022 challenge on perceptual image quality assessment.
\newblock In {\em CVPRW}, pages 951--967, 2022.

\bibitem{jinjin2020pipal}
Gu Jinjin, Cai Haoming, Chen Haoyu, Ye Xiaoxing, Jimmy~S Ren, and Dong Chao.
\newblock Pipal: a large-scale image quality assessment dataset for perceptual
  image restoration.
\newblock In {\em ECCV}, pages 633--651, 2020.

\bibitem{ke2021musiq}
Junjie Ke, Qifei Wang, Yilin Wang, Peyman Milanfar, and Feng Yang.
\newblock Musiq: Multi-scale image quality transformer.
\newblock In {\em ICCV}, pages 5148--5157, 2021.

\bibitem{liang2024vrt}
Jingyun Liang, Jiezhang Cao, Yuchen Fan, Kai Zhang, Rakesh Ranjan, Yawei Li,
  Radu Timofte, and Luc Van~Gool.
\newblock Vrt: A video restoration transformer.
\newblock {\em IEEE Trans. Image Process.}, 33:2171--2182, 2024.

\bibitem{liang2022recurrent}
Jingyun Liang, Yuchen Fan, Xiaoyu Xiang, Rakesh Ranjan, Eddy Ilg, Simon Green,
  Jiezhang Cao, Kai Zhang, Radu Timofte, and Luc~V Gool.
\newblock Recurrent video restoration transformer with guided deformable
  attention.
\newblock In {\em NeurIPS}, pages 378--393, 2022.

\bibitem{loshchilov2017decoupled}
Ilya Loshchilov and Frank Hutter.
\newblock Decoupled weight decay regularization.
\newblock {\em arXiv preprint arXiv:1711.05101}, 2017.

\bibitem{lu2025dpm}
Cheng Lu, Yuhao Zhou, Fan Bao, Jianfei Chen, Chongxuan Li, and Jun Zhu.
\newblock Dpm-solver++: Fast solver for guided sampling of diffusion
  probabilistic models.
\newblock {\em arXiv preprint arXiv:2211.01095}, 2022.

\bibitem{mittal2012making}
Anish Mittal, Rajiv Soundararajan, and Alan~C Bovik.
\newblock Making a “completely blind” image quality analyzer.
\newblock {\em IEEE Sign. Process. Letters}, 20(3):209--212, 2012.

\bibitem{nah2019ntire}
Seungjun Nah, Sungyong Baik, Seokil Hong, Gyeongsik Moon, Sanghyun Son, Radu
  Timofte, and Kyoung Mu~Lee.
\newblock Ntire 2019 challenge on video deblurring and super-resolution:
  Dataset and study.
\newblock In {\em CVPRW}, pages 0--0, 2019.

\bibitem{pan2021deep}
Jinshan Pan, Haoran Bai, Jiangxin Dong, Jiawei Zhang, and Jinhui Tang.
\newblock Deep blind video super-resolution.
\newblock In {\em ICCV}, pages 4811--4820, 2021.

\bibitem{pan2020cascaded}
Jinshan Pan, Haoran Bai, and Jinhui Tang.
\newblock Cascaded deep video deblurring using temporal sharpness prior.
\newblock In {\em CVPR}, pages 3043--3051, 2020.

\bibitem{peebles2023scalable}
William Peebles and Saining Xie.
\newblock Scalable diffusion models with transformers.
\newblock In {\em CVPR}, pages 4195--4205, 2023.

\bibitem{podell2023sdxl}
Dustin Podell, Zion English, Kyle Lacey, Andreas Blattmann, Tim Dockhorn, Jonas
  M{\"u}ller, Joe Penna, and Robin Rombach.
\newblock Sdxl: Improving latent diffusion models for high-resolution image
  synthesis.
\newblock {\em arXiv preprint arXiv:2307.01952}, 2023.

\bibitem{raffel2020exploring}
Colin Raffel, Noam Shazeer, Adam Roberts, Katherine Lee, Sharan Narang, Michael
  Matena, Yanqi Zhou, Wei Li, and Peter~J Liu.
\newblock Exploring the limits of transfer learning with a unified text-to-text
  transformer.
\newblock {\em J. Mach. Learn. Res.}, 21(140):1--67, 2020.

\bibitem{rombach2022high}
Robin Rombach, Andreas Blattmann, Dominik Lorenz, Patrick Esser, and Bj{\"o}rn
  Ommer.
\newblock High-resolution image synthesis with latent diffusion models.
\newblock In {\em CVPR}, pages 10684--10695, 2022.

\bibitem{stergiou2022adapool}
Alexandros Stergiou and Ronald Poppe.
\newblock Adapool: Exponential adaptive pooling for information-retaining
  downsampling.
\newblock {\em IEEE Trans. Image Process.}, 32:251--266, 2022.

\bibitem{su2017deep}
Shuochen Su, Mauricio Delbracio, Jue Wang, Guillermo Sapiro, Wolfgang Heidrich,
  and Oliver Wang.
\newblock Deep video deblurring for hand-held cameras.
\newblock In {\em CVPR}, pages 1279--1288, 2017.

\bibitem{tao2017detail}
Xin Tao, Hongyun Gao, Renjie Liao, Jue Wang, and Jiaya Jia.
\newblock Detail-revealing deep video super-resolution.
\newblock In {\em ICCV}, pages 4472--4480, 2017.

\bibitem{wang2023exploring}
Jianyi Wang, Kelvin~CK Chan, and Chen~Change Loy.
\newblock Exploring clip for assessing the look and feel of images.
\newblock In {\em AAAI}, pages 2555--2563, 2023.

\bibitem{wang2025seedvr2}
Jianyi Wang, Shanchuan Lin, Zhijie Lin, Yuxi Ren, Meng Wei, Zongsheng Yue,
  Shangchen Zhou, Hao Chen, Yang Zhao, Ceyuan Yang, et~al.
\newblock Seedvr2: One-step video restoration via diffusion adversarial
  post-training.
\newblock {\em arXiv preprint arXiv:2506.05301}, 2025.

\bibitem{wang2025seedvr}
Jianyi Wang, Zhijie Lin, Meng Wei, Yang Zhao, Ceyuan Yang, Chen~Change Loy, and
  Lu Jiang.
\newblock Seedvr: Seeding infinity in diffusion transformer towards generic
  video restoration.
\newblock In {\em CVPR}, pages 2161--2172, 2025.

\bibitem{wang2024exploiting}
Jianyi Wang, Zongsheng Yue, Shangchen Zhou, Kelvin~CK Chan, and Chen~Change
  Loy.
\newblock Exploiting diffusion prior for real-world image super-resolution.
\newblock {\em Int. J. Comput. Vis.}, 132(12):5929--5949, 2024.

\bibitem{wang2019edvr}
Xintao Wang, Kelvin~CK Chan, Ke Yu, Chao Dong, and Chen Change~Loy.
\newblock Edvr: Video restoration with enhanced deformable convolutional
  networks.
\newblock In {\em CVPRW}, pages 0--8, 2019.

\bibitem{wang2021real}
Xintao Wang, Liangbin Xie, Chao Dong, and Ying Shan.
\newblock Real-esrgan: Training real-world blind super-resolution with pure
  synthetic data.
\newblock In {\em CVPR}, pages 1905--1914, 2021.

\bibitem{wang2018esrgan}
Xintao Wang, Ke Yu, Shixiang Wu, Jinjin Gu, Yihao Liu, Chao Dong, Yu Qiao, and
  Chen Change~Loy.
\newblock Esrgan: Enhanced super-resolution generative adversarial networks.
\newblock In {\em ECCVW}, pages 0--0, 2018.

\bibitem{wu2023exploring}
Haoning Wu, Erli Zhang, Liang Liao, Chaofeng Chen, Jingwen Hou, Annan Wang,
  Wenxiu Sun, Qiong Yan, and Weisi Lin.
\newblock Exploring video quality assessment on user generated contents from
  aesthetic and technical perspectives.
\newblock In {\em ICCV}, pages 20144--20154, 2023.

\bibitem{xie2025star}
Rui Xie, Yinhong Liu, Penghao Zhou, Chen Zhao, Jun Zhou, Kai Zhang, Zhenyu
  Zhang, Jian Yang, Zhenheng Yang, and Ying Tai.
\newblock Star: Spatial-temporal augmentation with text-to-video models for
  real-world video super-resolution.
\newblock {\em arXiv preprint arXiv:2501.02976}, 2025.

\bibitem{xue2019video}
Tianfan Xue, Baian Chen, Jiajun Wu, Donglai Wei, and William~T Freeman.
\newblock Video enhancement with task-oriented flow.
\newblock {\em Int. J. Comput. Vis.}, 127(8):1106--1125, 2019.

\bibitem{yang2024motion}
Xi Yang, Chenhang He, Jianqi Ma, and Lei Zhang.
\newblock Motion-guided latent diffusion for temporally consistent real-world
  video super-resolution.
\newblock In {\em ECCV}, pages 224--242. Springer, 2024.

\bibitem{yang2024cogvideox}
Zhuoyi Yang, Jiayan Teng, Wendi Zheng, Ming Ding, Shiyu Huang, Jiazheng Xu,
  Yuanming Yang, Wenyi Hong, Xiaohan Zhang, Guanyu Feng, et~al.
\newblock Cogvideox: Text-to-video diffusion models with an expert transformer.
\newblock {\em arXiv preprint arXiv:2408.06072}, 2024.

\bibitem{yi2019progressive}
Peng Yi, Zhongyuan Wang, Kui Jiang, Junjun Jiang, and Jiayi Ma.
\newblock Progressive fusion video super-resolution network via exploiting
  non-local spatio-temporal correlations.
\newblock In {\em ICCV}, pages 3106--3115, 2019.

\bibitem{yu2024scaling}
Fanghua Yu, Jinjin Gu, Zheyuan Li, Jinfan Hu, Xiangtao Kong, Xintao Wang,
  Jingwen He, Yu Qiao, and Chao Dong.
\newblock Scaling up to excellence: Practicing model scaling for
  photo-realistic image restoration in the wild.
\newblock In {\em CVPR}, pages 25669--25680, 2024.

\bibitem{zhang2021designing}
Kai Zhang, Jingyun Liang, Luc Van~Gool, and Radu Timofte.
\newblock Designing a practical degradation model for deep blind image
  super-resolution.
\newblock In {\em CVPR}, pages 4791--4800, 2021.

\bibitem{zhang2023adding}
Lvmin Zhang, Anyi Rao, and Maneesh Agrawala.
\newblock Adding conditional control to text-to-image diffusion models.
\newblock In {\em ICCV}, pages 3836--3847, 2023.

\bibitem{zhang2018unreasonable}
Richard Zhang, Phillip Isola, Alexei~A Efros, Eli Shechtman, and Oliver Wang.
\newblock The unreasonable effectiveness of deep features as a perceptual
  metric.
\newblock In {\em CVPR}, pages 586--595, 2018.

\bibitem{zhang2023md}
Zicheng Zhang, Wei Wu, Wei Sun, Danyang Tu, Wei Lu, Xiongkuo Min, Ying Chen,
  and Guangtao Zhai.
\newblock Md-vqa: Multi-dimensional quality assessment for ugc live videos.
\newblock In {\em CVPR}, pages 1746--1755, 2023.

\bibitem{zhou2024upscale}
Shangchen Zhou, Peiqing Yang, Jianyi Wang, Yihang Luo, and Chen~Change Loy.
\newblock Upscale-a-video: Temporal-consistent diffusion model for real-world
  video super-resolution.
\newblock In {\em CVPR}, pages 2535--2545, 2024.

\end{thebibliography}
}

\clearpage

\appendix
\section{Appendix}

We present additional analysis of the proposed method here.
First, we discuss the trade-off between fidelity and quality in the proposed Vivid-VR.
Second, we further analyze the effect of the number of  training videos generated by the proposed concept distillation strategy.
Furthermore, we discuss the effect of the number of ControlNet blocks.
Finally, we present more visualization results.

\begin{figure*}[h]
	\centering
	\includegraphics[width=0.88\linewidth]{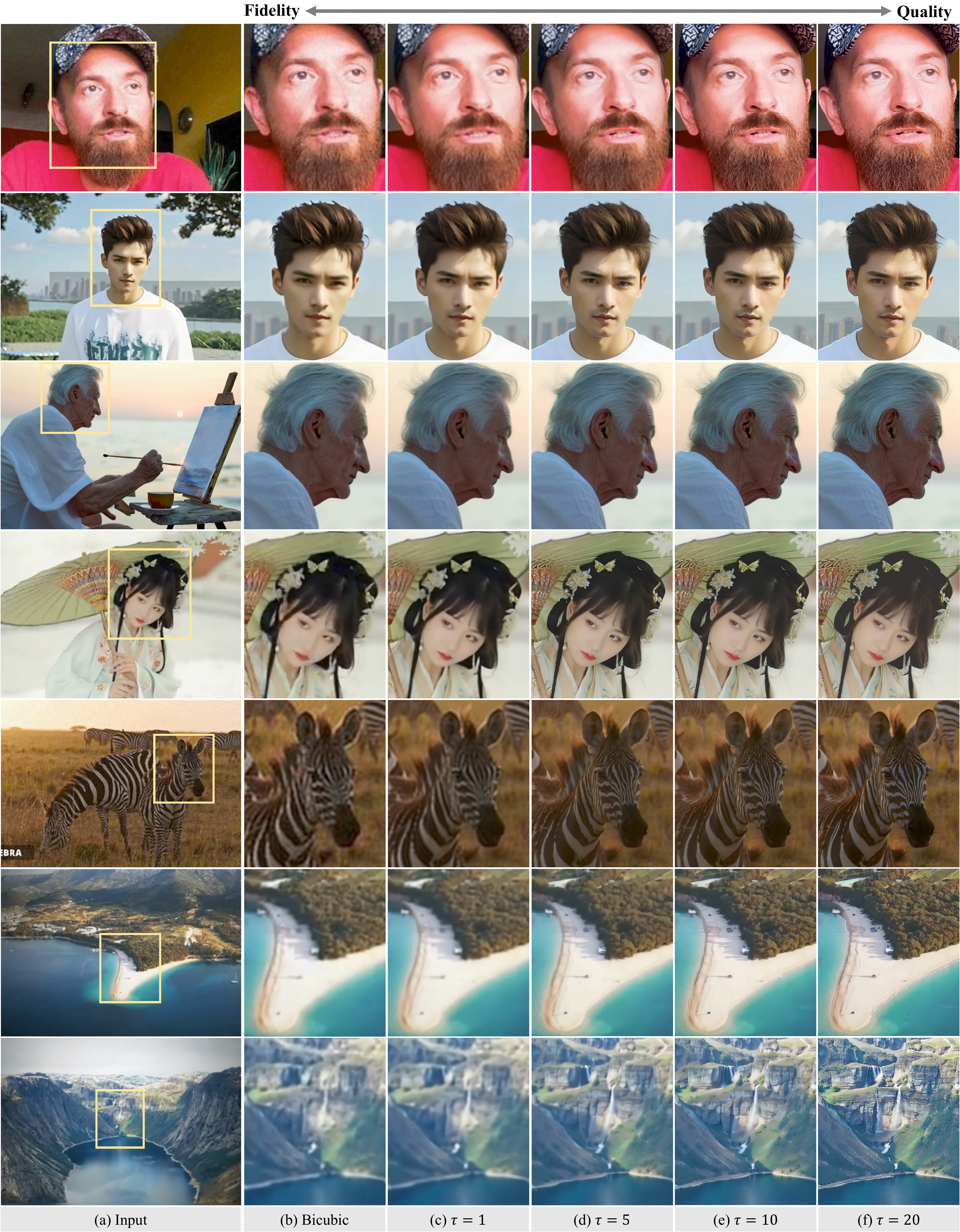}
	\vspace{1mm}
	\caption{%
		Trade-off between fidelity and quality. Higher guidance coefficient $\tau$ in the \textit{Restoration-Guided Sampling} yield more realistic results, while lower $\tau$ preserve greater fidelity to the original input content. (\textbf{Zoom-in for best view})
	}
	\label{fig:supp-restoration-guidance}
\end{figure*}

\noindent\textbf{Trade-off between Fidelity and Quality.}
As~\cite{yu2024scaling} points out, powerful generative prior is a double-edged sword, as excessive generative capacity may in turn affect the fidelity of the restored video.
To address this, we introduce \textit{Restoration-Guided Sampling} into Vivid-VR's inference sampling process to balance the quality and fidelity:
\begin{equation}
	\setlength{\abovedisplayskip}{3pt}  
	\setlength{\belowdisplayskip}{3pt}  
	\hat{x}^{est}_t = x^{est}_t+(\frac{t}{T} )^\tau (x^{lq}-x^{est}_t),
	\label{eq:restoration-guided}
\end{equation}
where $x^{est}$ is the denoised latent at time step $t$, and $x^{lq}$ is the original input latent; $T$ denotes the total number of denoising steps; $\tau$ is the guidance coefficient; $\hat{x}^{est}$ is the output latent after the restoration-guided sampling.
Figure~\ref{fig:supp-restoration-guidance} demonstrates this trade-off: higher guidance coefficient $\tau$ yield more realistic results, while lower $\tau$ preserve greater fidelity to the original input content.

\noindent\textbf{Effect of the Number of  Generated Training Videos.}
As mentioned in the main paper, the proposed method employs the concept distillation strategy to generate $100K$ videos for training.
A natural question arises: does the number of generated training videos impact restoration performance?
To investigate this, we conducted the ablation studies here. 
Table~\ref{tab:supp-ablation-cd} demonstrates that increasing the number of generated training videos from $0$ to $100K$ yields significant performance gains, while expanding from $100K$ to $150K$ shows diminishing returns. 
Considering the cost of generating training videos, we therefore adopt $100K$ generated videos as our standard configuration.
Furthermore, we verified that relying solely on generated training data (without source videos) leads to suboptimal results.
This occurs because the T2V base model's outputs contain inherent imperfections.
Training exclusively on such data ultimately compromises model performance.


\begin{table}[h]
	\caption{Effect of the number of training videos generated by the proposed concept distillation.}
	\vspace{-2mm}
	\renewcommand\arraystretch{1.1}
	\center
	\scriptsize
	\setlength{\tabcolsep}{5.5pt}
	\begin{tabular}{c|c|c|cccc}
		\toprule
		\multirow{2}{*}{} & \multicolumn{2}{c|}{{Training Videos}} & \multirow{2}{*}{NIQE} & \multirow{2}{*}{MUSIQ} & \multirow{2}{*}{CLIP-IQA} & \multirow{2}{*}{DOVER} \\ \cline{2-3}
		&  \# Source   &  \# Generated    &      &       &          &       \\ \hline
		(a)     & 500K  & 0             & 5.364                 & 57.36    & 0.363    & 12.99    \\
		(b)     & 500K  & 50K             & 4.562                 & 63.00    & 0.408    & 13.46    \\
		(c)     & 500K  & 100K             & 4.361                 & \textbf{67.61}     & \textbf{0.450}     & 14.46    \\
		(d)     & 500K  & 150K             &  \textbf{4.292}                 &67.19    & 0.443    & \textbf{14.51}    \\
		(e)     & 0  & 150K           & 5.652                 & 53.77    & 0.377    & 11.63    \\
		\bottomrule
	\end{tabular}
	\label{tab:supp-ablation-cd}
\end{table}

\noindent\textbf{Effect of the Number of ControlNet Blocks.}
To reduce the parameter count, we employ $N/7$ DiT blocks in ControlNet.
In~\cite{deng2025acquire}, only one block is used, and all connectors share the same control feature.
To further investigate whether $N/7$ DiT blocks are indeed necessary, we set the number of blocks to $1$ and retrain using the same settings.
The results in Table~\ref{tab:supp-controlnet-block} show that using only one block does not perform well.

\begin{table}[h]
\caption{Effect of the number of ControlNet blocks.}
\vspace{-2mm}
\renewcommand\arraystretch{1.1}
\center
\scriptsize
	\begin{tabular}{c|cccc}
	\toprule
	 & NIQE & MUSIQ & CLIP-IQA & DOVER  \\ \hline
	$1$ block    & 4.855   &  66.85   &  0.442   &  14.17   \\
	$N/7$ blocks (Vivid-VR)  & \textbf{4.361}          & \textbf{67.61}     & \textbf{0.450}     & \textbf{14.46}    \\
	\bottomrule
\end{tabular}
\label{tab:supp-controlnet-block}
\end{table}

\noindent\textbf{More Visualization Results.}
In the main paper, we have demonstrated that the proposed Vivid-VR achieves state-of-the-art performance.
Furthermore, we provide more visual comparisons on the project homepage (\url{https://csbhr.github.io/projects/vivid-vr/}), where Vivid-VR demonstrates superior structural clarity, texture richness, and visual vividness.

\end{document}